\documentclass[letterpaper, 10 pt, conference]{ieeeconf}  

\IEEEoverridecommandlockouts                              

\title{\LARGE \bf
Error Diagnosis of Deep Monocular Depth Estimation Models
}

\author{Jagpreet Chawla$^{1}$, Nikhil Thakurdesai$^{1}$, Anuj Godase$^{1}$, Md Reza$^{1}$, David Crandall$^{1}$ and Soon-Heung Jung$^{2}$
\thanks{$^{1}$Luddy School of Informatics, Computing, and Engineering, Indiana University, Bloomington, IN 47408, USA
        {\tt\small \{jchawla, nthakurd, abgodase, mdreza, djcran\}@iu.edu}}%
\thanks{$^{2}$Electronics and Telecommunications Research Institute, Daejeon 34129, South Korea
        {\tt\small zeroone@etri.re.kr}}%
\thanks{This work was supported by Electronics and Telecommunications Research Institute (ETRI) grant funded by the Korean government (21ZH1200, The research of the fundamental media·contents technologies for hyper-realistic media space).}        
}

\usepackage{booktabs}
\usepackage{graphicx}
\usepackage{mathtools}
\usepackage{color}

\usepackage{multirow}
\usepackage{subcaption}
\usepackage{graphicx}

\begin{document}

\maketitle
\thispagestyle{empty}
\pagestyle{empty}

\begin{abstract}

Estimating depth from a monocular image is an ill-posed problem: when the camera projects a 3D scene onto a 2D plane, depth information is inherently and permanently lost. Nevertheless, recent work has shown impressive results in estimating 3D structure from 2D images using deep learning. In this paper, we put on an introspective hat and analyze state-of-the-art monocular depth estimation models in indoor scenes to understand these models' limitations and error patterns. To address errors in depth estimation, we introduce a novel \textit{Depth Error Detection Network (DEDN)} that spatially identifies erroneous depth predictions in the monocular depth estimation models. By experimenting with multiple state-of-the-art monocular indoor depth estimation models on multiple datasets, we show that our proposed depth error detection network can identify a significant number of errors in the predicted depth maps. Our module is flexible and can be readily plugged into any monocular depth prediction network to help diagnose its results. Additionally, we propose a simple yet effective \textit{Depth Error Correction Network (DECN)} that iteratively corrects  errors based on our initial error diagnosis.


\end{abstract}

\section{INTRODUCTION}



%

Monocular depth estimation is an important problem in  robotics and computer vision. Depth maps can be used to understand the 3D structure and relative positions of objects in a scene for applications including autonomous driving~\cite{Wang_2019_CVPR}, visual odometry~\cite{zhan_icra2020,yang2018_dvso}, augmented reality~\cite{lee_ICAT11}, sensor fusion~\cite{Fcil2017SingleViewAM},
and many others. Estimating depth from a monocular image is an inherently ill-posed
problem, since 3D information is irretrievably lost when the camera
projects to a 2D image.

Nevertheless, visual cues such as shadows, highlights, defocus, and silhouettes can be exploited to approximately recover the depth map of a scene. Machine learning-based approaches such as Make3D~\cite{saxena_nips06}, and more recent techniques based
on deep learning~\cite{eigen_iccv15,laina_3dv2016}, have shown significant promise. These techniques take a variety of approaches. For example, instead of directly estimating depth, BTS~\cite{lee_arxiv19} estimates the parameters of local planes at
various scales. The model is trained using only ground truth depth, as the local plane parameters are learned implicitly by the network. 
PlaneRCNN~\cite{Liu_CVPR19}, another  state-of-the-art
technique, estimates planar surfaces in addition to estimating depth for non-planar
areas. The final depth map is then produced by combining these two
types of outputs.

Of course, these papers (and the countless others on monocular depth estimation)  
present quantitative results that characterize measures of
error with respect to ground truth. However, these quantitative error metrics
can be surprisingly opaque, making it difficult to choose among algorithms for any given application.
For example, multiple algorithms could yield results with exactly the same mean squared depth error, but
the error patterns could be completely different: one could have all depths under or over-estimated
by some offset, another could have most depth values exactly accurate but with a few extreme
outliers, while another could accurately estimate the depth of object surfaces but give inaccurate
estimates for object boundaries. Despite having the same quantitative errors, these three
algorithms would have very different performance in a real-world application.

\begin{figure}[t] 
      \centering
      \includegraphics[width=0.15\textwidth]{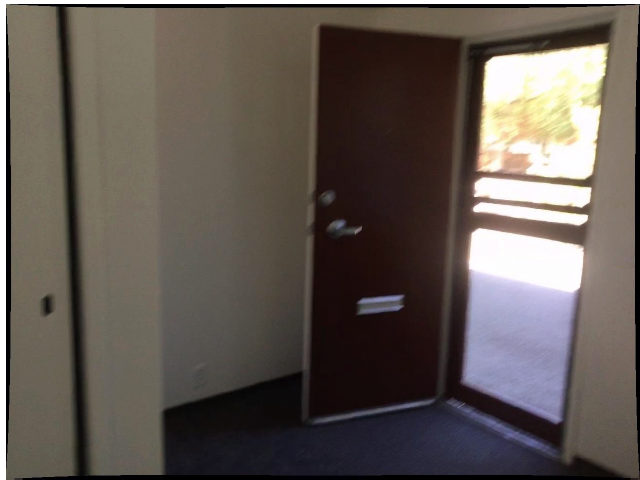}
      \includegraphics[width=0.15\textwidth]{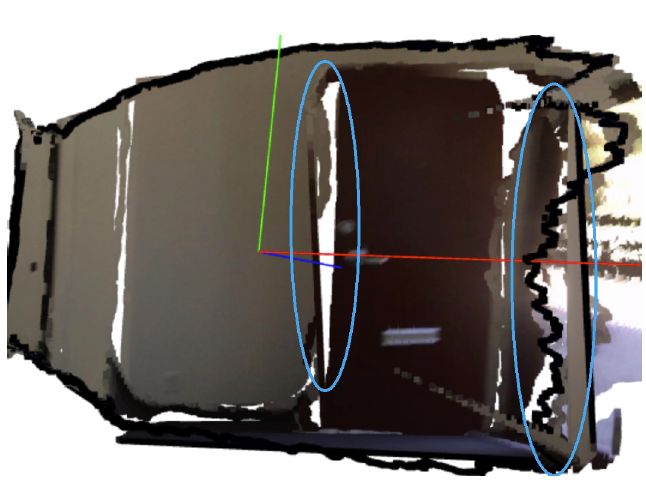}
      \includegraphics[width=0.15\textwidth]{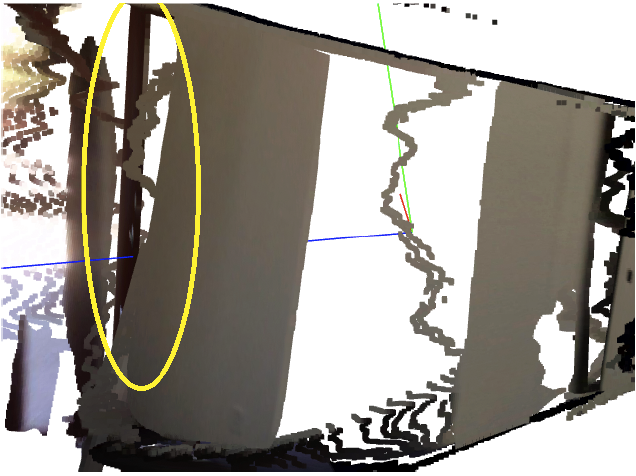}
      \vspace{1mm}
       \caption{Sample errors in predicted depth using Plane-RCNN~\cite{Liu_CVPR19}. From left to right: A sample input, and two different views of the generated 3D output.  
            The blue area denotes segmentation issues around the boundaries of the plane, while the yellow area shows planes that are supposed to be adjacent and connected  (best viewed in color).}
      \label{fig:vis_depth_error_motivation}
\end{figure}

In this paper, we propose a technique to analyze  methods in monocular depth estimation and to \textit{spatially identify and characterize} likely errors in their output. We evaluate this technique on three diverse approaches to monocular 3D estimation, PlaneRCNN~\cite{lee_arxiv19}, Eigen et al.~\cite{eigen_iccv15}, and  BTS~\cite{lee_arxiv19}, and experiment on two different datasets, NYUDv2~\cite{Silberman_ECCV12} and ScanNet~\cite{scannet_cvpr2017}. We find that our error diagnostic tools can \textit{identify erroneously predicted depth locations} in monocular depth estimation methods. Additionally, we  propose a simple yet effective iterative method to correct the likely errors, showing that it can improve the depth map estimates. Figure~\ref{fig:vis_depth_error_motivation} demonstrates some of the errors when the same scene is visualized from different viewpoints. 

More specifically, we make the following contributions.
    First, we introduce a Depth Error Detection Network (DEDN) to help diagnose model errors. Our method can  locate pixels  with likely erroneous depth estimates. Additionally, we propose numerical measures  to quantify properties of incorrectly predicted depth locations.
    Second, we evaluate  DEDN on two  datasets (NYUDv2 and ScanNet) in single-view and  multi-view settings and using different depth prediction methods.
   We show it is generic and  can be applied to any depth predictor.
    Third, we introduce a Depth Error Correction Network (DECN) to iteratively correct  errors  detected by DEDN.

\section{Related work}

\subsection*{Introspective Capability of Machine Learning Models}

Understanding what a model does not know is a critical part of many
applications.
Grimmett et al.~\cite{grimmett_icra13,grimmet_ijrr16}
raised concerns about the limitations of existing metrics (e.g., {precision} and
{recall}) for evaluating classification. They showed that
classifiers like {SVMs} and {LogitBoost}  are
overconfident about their predictions. 
For high-stakes
applications like autonomous driving, the authors emphasized the need for an
introspective capability  and  introduced entropy
measurement-based uncertainty estimates during classification. With
this novel notion of the introspective capability of a classifier,
they demonstrated that the Gaussian process classifier is better
suited to some decision-making robotic systems.
%
Berczi et al.~\cite{berczi_icra15} reached a similar conclusion with  Gaussian processes  for a robotic terrain assessment system.

More recently, Gal et al.~\cite{gal2016dropout} proposed Monte Carlo dropout for Bayesian approximation to model uncertainties in deep learning. Lakshminarayanan et al.~\cite{lakshminarayanan2016simple} suggested deep ensemble-based uncertainty estimates and demonstrated their    value over Bayesian approximation.
Kendall et al.~\cite{kendall_nips17} identified different types of uncertainties that could arise during decision making, and explicitly encoded them into a Bayesian deep learning model. They demonstrated that this novel model is capable of identifying  uncertainties for monocular depth estimation. The first type of uncertainty, \textit{aleatoric uncertainty}, is inherent to the observations, such as
depth for a distant object  or at occlusion boundaries.
The second type, \textit{epistemic uncertainty}, is inherent
to the model, e.g., the uncertainty of model parameters, and can be resolved through better models or more training data. 
More recently, Posetels et al.~\cite{postels2019sampling} proposed a sampling-free strategy for estimating the {epistemic uncertainty}.


\subsection*{Error Diagnostic Measures and Metrics}
Error diagnostics have been explored for various image understanding tasks~\cite{boyla_eccv20,hoiem_eccv12} including  monocular depth prediction~\cite{cadena_iros16}.  Boyla et al.~\cite{boyla_eccv20} introduced a tool called TIDE for identifying different sources of errors in object detection and segmentation.
Cadena et al.~\cite{cadena_iros16} analyzed the limitations of the existing evaluation metrics --- e.g., {mean absolute error, root mean square error, etc.} --- to characterize depth prediction performance. They also proposed a new measure that addresses some deficiencies in  earlier metrics. 
Hekmatian et al.~\cite{hekmatian2019confnet} address depth completion from  sparse point-clouds of LiDAR sensors. 
Their error map prediction module, which is explicitly designed to model depth  errors, is jointly trained  with the depth map predictor. In contrast, we propose two error diagnostic modules to identify and correct errors in sequential stages, decoupled from the existing depth prediction network. Also, while~\cite{hekmatian2019confnet} aimed to produce dense depth maps from sparse input point clouds, 
our  method is designed to spatially quantify the errors produced by an existing depth prediction network first, and then to provide a mechanism to improve the predicted depth map. We evaluate  on three different depth prediction models and two datasets.

\section{Method}
We propose error diagnosis methods for the  depth maps produced by deep neural network-based techniques. First, we propose an error diagnostic method -- \textit{Depth Error Detection Network (DEDN)} -- that can identify locations of erroneous depth predictions in the output  of DNN-based depth estimation models. We devised two types of DEDNs to achieve this goal, one for single-view images (Sec~\ref{sec:single-view-error-detect-network}) and another for multiple views (Sec~\ref{sec:multi-view-error-detect-network}).
We also introduce a technique for correcting the detected errors (Sec~\ref{sec:error-correct-network}).

\subsection{Depth Error Detection Network (DEDN)}
\label{sec:error-detect-network}
Our DEDNs receive the output of a depth
estimation model and try to predict which pixels have incorrect
estimates. More precisely, our first DEDN
receives the predicted depth map, along with its corresponding RGB pixel value,
from an existing method as input pair, and then quantifies the
degree of inconsistencies in that predicted depth map. We refer to this as
our \textit{single-view DEDN}
(Figure~\ref{fig:single-view-error-detection-network}) since it
focuses solely on the input frame without exploiting error
patterns around surrounding frames. Our second error detection
network -- \textit{multi-view DEDN}
(Figure~\ref{fig:multi-view-error-detection-network}) -- exploits
additional adjacent frames for better error diagnosis.


\subsubsection{Single-view DEDN}
\label{sec:single-view-error-detect-network}

\begin{figure}[t]
    \centering
    \includegraphics[width=8cm]{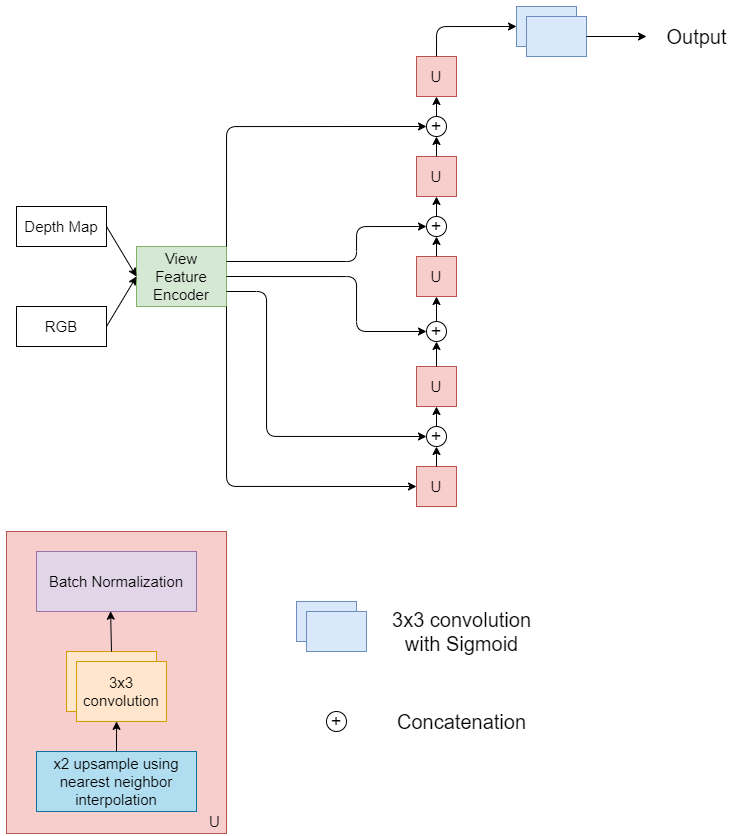}
    \caption{Single-view Depth Error Detection Network (DEDN),  consisting of a View Feature Encoder (Figure \ref{fig:view-feature-encoder}) that takes the Depth Map and  RGB image as input. The U blocks denote  Upsampling. 
     }
    \label{fig:single-view-error-detection-network}
\end{figure}

\begin{figure}[t]
    \centering
    \includegraphics[width=7cm]{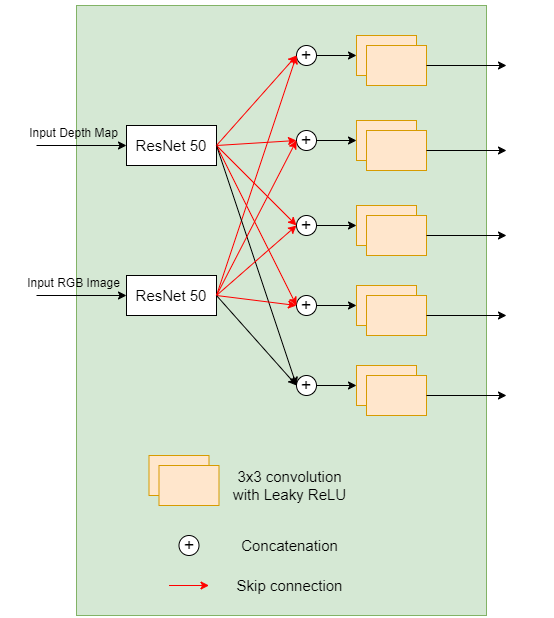}
    \caption{Our View Feature Encoder consists of two ResNet-50's, for the Input Depth Map and the Input RGB Image. The output feature maps are concatenated together. 
     }
    \label{fig:view-feature-encoder}
\end{figure}



We designed an encoder-decoder   network architecture for
detecting depth errors. 
Our encoder  can be thought of as a feature
extractor acting on an input pair -- a predicted depth map (from an
existing depth estimation method) and its corresponding RGB image. The
decoder then assumes the role of an error predictor that formulates
its task as per-pixel error classification. The low-resolution
feature maps produced by the encoder need to be upsampled. Inspired
by U-Net~\cite{unet_miccai15}, the decoder module uses skip
connections from the encoder at each level to aid with upsampling as
shown in Figure~\ref{fig:single-view-error-detection-network}.


\subsubsection*{View Feature Encoder}
As shown in Fig. \ref{fig:view-feature-encoder}, the View Feature Encoder takes a Predicted Depth Map along with a corresponding RGB image as input. We call this the ``view" feature encoder as this same encoder can take multiple views as input. For the single-view case, we just have one view, while for the multi-view case, we can use this same module for mutliple views \textit{without adding} to the complexity of our model (in terms of number of network parameters).  We  discuss  the multi-view case in more detail in Section \ref{sec:multi-view-error-detect-network}.
We use ResNet50v2~\cite{He2016DeepRL}  pre-trained on Imagenet \cite{deng2009imagenet} as an encoder for
both types of inputs -- predicted depth map and RGB -- although 
other network architectures could be used. 
Using a pre-trained network has the significant
advantage that the encoder already produces some useful features, 
providing a better starting point. Moreover, using an encoder trained
on a different dataset also promotes generalization. 
from depth input as well.


Since we want to find features from both the RGB image and
the predicted depth map and then detect discrepancies to find possible depth errors, ideally we would want a similar set of features.
To achieve this goal, we performed an extra pre-training step for our
depth encoder leveraging the idea of \textit{cross modal distillation}
to transfer knowledge from one image modality to
another~\cite{distillation_cvpr16}. Past 
work~\cite{distillation_cvpr16} showed that transfer of supervision from
one modality (e.g., RGB) to another results in significant improvement
in downstream tasks that use the second modality (e.g, depth).
By leveraging the concept of cross-modal distillation, we take
features of a network trained in one modality and train the second
network to produce the same features given the corresponding paired
image in a different modality. More precisely, to pre-train
the depth encoder for extracting features from depth input, we first
trained it to produce the same set of features as a pre-trained RGB
encoder. The depth encoder and RGB encoder  have the same architecture except for a different number of input channels. Using mean
squared error as a loss function, we trained the encoder module on
ScanNet~\cite{scannet_cvpr2017} as it contains a large set of paired
RGB and depth images for all of our experiments. Once we trained our depth encoder, we used it as
pre-trained encoders to train the complete depth error detection
model.

\subsubsection*{Decoder}
The decoder consists of several convolutions and upsampling steps,
with skip connections from encoders. At each step, the output of two
encoders are concatenated,  followed by a 2D 3x3
convolution. This is concatenated with the output of the previous
decoding step followed by upsampling with  nearest neighbor
interpolation, 2D 3x3 convolution, and batch normalization. All
convolutions use leaky ReLU activation, except for the final output
which is a 2D 3x3 convolution with sigmoid activation. See Figure~\ref{fig:single-view-error-detection-network}.

\subsubsection*{Loss function}
\label{sec:under_over_correct_classes}

We formulated the error prediction as a per-pixel classification task,
where the goal is to categorize the predicted depth estimations into
correct, over-estimated, or under-estimated.
To assign each pixel to a category, we check
the absolute difference between the estimated depth $\mathbf{D}$ and
the ground truth depth $\mathbf{D^*}$, as follows:
\begin{itemize}
\item \textbf{Correct} if the estimate is within a threshold $t$ of the ground truth, 
$ \lvert d_{i,j} - d_{i,j}^*\rvert \leq t $.

\item \textbf{Under-estimate} if the difference is more than $t$ and the predicted depth value is less than the ground truth, 
$ d_{i,j} - d_{i,j}^* < -t $.

\item \textbf{Over-estimate} if the difference is more than $ t $ and the predicted depth value is more than the ground truth,
$ d_{i,j} - d_{i,j}^*> t $.
\end{itemize}


%

\noindent
Then the cross-entropy loss of our error classification is,
\begin{equation*}
    L = - \dfrac{1}{N} \sum_{i=1}^{N} \sum_{j=1}^{3} m_{i,j} \cdot c_j \cdot y_{i,j} \cdot \log (\bar y_{i,j})
\end{equation*}
where $ N $ is the total number of pixels in the input image,
$ m_{i,j} $ is the binary mask for each pixel,
$ c_j $ is the weight assigned to that particular class, $ \bar
y_{i,j} $ is the probability that the estimated depth for pixel $ i $
belongs to class $ j $ as calculated by the DEDN
and $ y_{i,j} $ is the ground truth probability for pixel $ i $
belonging to class $ j $.


The training dataset of our DEDN model is inherently
imbalanced and the degree of imbalance depends on the depth prediction -- e.g., a very good depth predictor has many more correct pixels than incorrect ones. 
We handle the imbalance by using a weighted loss where the weight of each
pixel is inversely proportional to the number of samples of that
pixel's class in the dataset, so that the class with the greater
amount of samples is weighted less and vice versa.
Pixels with missing depth are ignored in the loss
computation by assigning them a weight of zero to ensure that they do
not negatively affect the training of our DEDN. 


\subsubsection{Multi-view DEDN}
\label{sec:multi-view-error-detect-network}

\begin{figure}[t]
    \centering
    \includegraphics[width=8cm]{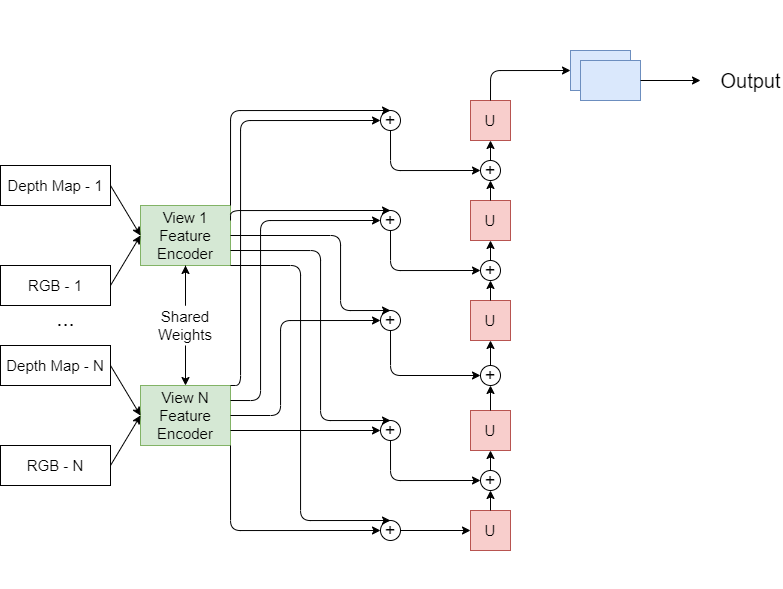}
    \caption{Multi-view Depth Error Detection Network (DEDN) is based on single-view DEDN 
    but with multiple View Feature Encoders. Moreover, there is an extra concatenation step to gather information from multiple views before doing the upsampling. The ``U" blocks denote  Upsampling blocks.
    }
    \label{fig:multi-view-error-detection-network}
\end{figure}

We follow a similar architecture   for the {multi-view DEDN}. Our View Feature Encoder module allows us to take any number of views as input  (Figure~\ref{fig:multi-view-error-detection-network}). The weights of the encoder are shared between all the views, so increasing views does not add to the model's complexity.
We use two views by taking an  adjacent frame in addition
the frame for which we are detecting depth errors. Our encoder can be thought of as a Siamese network with two branches, with each frame fed into its own branch.
Our intuition for incorporating an additional frame is to
provide additional cues which could help
identify errors which are inherently challenging to address from
a single view, such as occlusion, clutter, or other appearance
variations. 
In the two-view case, given the left RGB image, left depth map, right RGB image,  right depth map, the output embeddings from the Siamese network are then concatenated channel-wise and passed through a 3x3 convolution before sending them to the decoder (Figure~\ref{fig:multi-view-error-detection-network}). The same concatenation technique can be used for an arbitrary number of views. Again, we used
the same loss function as 
{single-view DEDN} to classify errors into categories
\textit{under-estimated} and \textit{over-estimated}.

\subsection{Depth Error Correction Network (DECN)}
\label{sec:error-correct-network}


\begin{figure}[t]
    \centering
    \includegraphics[width=8cm]{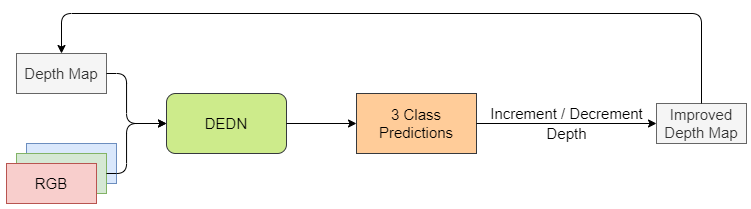}
    \caption{Depth Error Correction Network (DECN). We can plug in the Single-view  or Multi-view version of DEDN.}
    \label{fig:iterative_improvement}
\end{figure}

One application of DEDN is to refine the depth map by trying to correct
the errors it has identified. We can do this by incrementally
increasing depths of pixels that are predicted to be under-estimated,
and decreasing the depths of those predicted to be over-estimated.
Once each pixel is classified using the DEDN, we only consider pixels predicted as an error with
high confidence (greater than 0.7) and adjust the depth values by
incrementing or decrementing them based on the direction, by a fixed
small amount (0.01 meters in our experiments). These adjustments yield a new depth map, and the process
can be repeated multiple times as shown in
Figure~\ref{fig:iterative_improvement}. We can plug in either version of our DEDN -- Single-view or Multi-view.



\section{Experiments}

We experimented with three deep neural network-based monocular depth
estimation methods using our  error diagnostic modules on two different indoor datasets.

\subsection{Depth Prediction Models}

\textbf{Eigen}~\cite{eigen_iccv15} introduced an early DNN 
to estimate depth using a stack of two neural networks,
one to estimate a coarse depth map using global context, and another to refine this prediction locally. 
\textbf{Plane RCNN}~\cite{Liu_CVPR19} detects planes and their parameters along
with a depth map. The final depth map can be produced by combining per
pixel depth for non-planar regions and plane parameters for detected
planar regions.
\textbf{BTS} (``From Big to Small'')~\cite{lee_arxiv19} employs a novel local
planar guidance layer to produce depth cues at various scales using
a local planar assumption. The final depth map is estimated by using
these depth cues as input to final convolution layers.

\subsection{Datasets}

\textbf{NYUv2}~\cite{Silberman_ECCV12} dataset contains over 120,000 RGB-D images gathered from 464 scenes using a Microsoft Kinect. We use the official train-test split with 249 scenes for training and 215 scenes for testing. After aligning and synchronizing the RGB and depth data, we have 24,231 images for training and 654 images for testing.
%
%
Two common artifacts from Kinect-collected datasets are holes with missing depth
information and edge erosion~\cite{3dcameraLimitations}. 
%
\textbf{ScanNet}~\cite{scannet_cvpr2017} is a large RGB-D dataset containing 1,513 indoor scenes and 2.5 million views. It contains 3D camera pose information, surface reconstruction, and semantic segmentation. ScanNet used a Structure sensor~\cite{structure}, which is a portable 3D sensor designed similar to Microsoft Kinect v1. 
ScanNet is also affected
by artifacts such as holes with missing depth
information and edge erosion~\cite{3dcameraLimitations}.
We conduct evaluations on the respective datasets on which the model was trained. We do not explore cross-dataset evaluation or domain adaptation.

\begin{figure}
  \begin{subfigure}{1.0\columnwidth}
  \includegraphics[width=\textwidth]{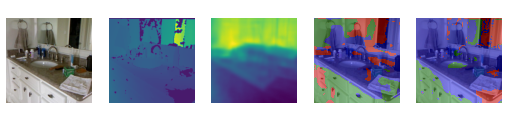}
  \end{subfigure}
  \hfill
  \begin{subfigure}{1.0\columnwidth}
  \includegraphics[width=\textwidth]{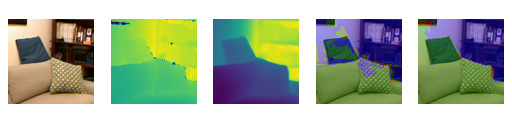}
  \end{subfigure} 
  \begin{subfigure}{1.0\columnwidth} 
  \includegraphics[width=\textwidth]{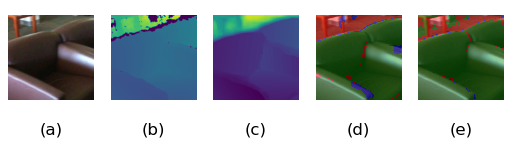} 
  \end{subfigure}  
  \caption{Sample visualizations of our single-view error detection. Each row denotes error diagnosis on (top to bottom) Eigen~\cite{eigen_iccv15}, BTS~\cite{lee_arxiv19}, and Plane-RCNN~\cite{Liu_CVPR19}. 
  The columns denote: (a)  RGB input, (b) ground truth depth, (c) predicted depth from the model, (d) errors in the output as
  predicted by our model,  (e) actual errors
  in the output  compared to ground truth. For (d) and (e), 
  \textit{red} means under-estimated, 
    \textit{green} means correct, and \textit{blue} means over-estimated.}
    
  \end{figure}

\subsection{Detecting Errors from the Depth Prediction Models}

In order to understand how well our model has learned to detect pixels
which have incorrect depth predictions (either under-estimated or
over-estimated), we use two metrics -- precision and recall --
for both the under-estimated and over-estimated classes. We would
ideally like a high precision with a reasonable recall  (e.g., greater than 50\%). This is because we do not want the network to
output too many false positives (low precison) as this would hurt the
performance of our DECN. At the same time, we do not
want many  false negatives (low recall) either, as this would mean
that we are missing out on correcting those pixels.


\subsubsection*{Random Baseline}
To the best of our knowledge, our error diagnosis work (DEDN) is first of its kind. We formulate the problem of error diagnosis of depth prediction into a three-class classification problem: i) under-estimated, ii) correct, and iii) over-estimated. No prior work has tried to classify errors in similar manner, which makes it difficult to compare our results with others. Instead we compare error diagnostic performance of our DEDN against a random baseline.
\begin{table}[t]
\resizebox{\columnwidth}{!}{
\centering
\begin{tabular}{@{}lcccc@{}} \toprule
\textbf{Model} &
  \textbf{\begin{tabular}[c]{@{}c@{}}Under-\\estimated \\ Precision ($\uparrow$)\end{tabular}} &
  \textbf{\begin{tabular}[c]{@{}c@{}}Over-\\estimated \\ Precision ($\uparrow$) \end{tabular}} &
  \textbf{\begin{tabular}[c]{@{}c@{}}Under-\\estimated \\ Recall ($\uparrow$) \end{tabular}} &
  \textbf{\begin{tabular}[c]{@{}c@{}}Over-\\estimated \\ Recall ($\uparrow$) \end{tabular}} \\ \midrule
\begin{tabular}[c]{@{}c@{}} Random Baseline \end{tabular} & 0.3152 & 0.2775 & \underline{0.3707}  & 0.2806  \\
Eigen~\cite{eigen_iccv15} single-view  & \underline{0.4965} & \underline{0.4199} & 0.3585 & \underline{0.4331} \\
Eigen~\cite{eigen_iccv15} multi-view    & \textbf{0.8862} & \textbf{0.8246} & \textbf{0.9292} & \textbf{0.7235} \\ \\
Random Baseline & 0.5045 & 0.1427 & 0.5091  & 0.1495  \\
Plane-RCNN~\cite{Liu_CVPR19} single-view                                                 & \underline{0.5910}  & \underline{0.2290}  & \underline{0.5250}  & \underline{0.2760}  \\ 
Plane-RCNN~\cite{Liu_CVPR19} multi-view & \textbf{0.6824} & \textbf{0.5538} & \textbf{0.8509} & \textbf{0.3794} \\ \\
Random Baseline & 0.0892 & 0.2011 & \textbf{0.1730}  & 0.2691  \\
BTS~\cite{lee_arxiv19} single-view                                                     & \underline{0.1919} & \underline{0.2431} & 0.0141 & \underline{0.2925} \\
BTS~\cite{lee_arxiv19} multi-view     & \textbf{0.5235} & \textbf{0.6934} & \underline{0.1484} & \textbf{0.7177} \\ \bottomrule \\
\end{tabular}
}
\caption{Error detection results. Higher  numbers are better.
Under-/over-estimated pixels have estimated depths
less/greater than the ground truth, respectively.
%
Numbers in \textbf{bold} are best  and  \underline{underline} are second best. 
}
\label{table:dedn_single_view}
\end{table}

Each model (Eigen, BTS or Plane-RCNN) has its own distribution for the three classes. 
For example, BTS has 56\%  correct, 17\% under-estimated, and 27\% over-estimated pixels, while Eigen has 35\%  correct, 37\% under-estimated, and 28\% over-estimated. To make comparisons fair, we use a random baseline for each model, where pixels are randomly assigned to the classes according to the distribution of model we are using. To calculate the precision and recall for the random baseline, we generate 10 random class predictions each of size 224 x 224. This gives about a half million pixel samples (10 x 224 x 224).

\subsubsection*{Single-view Results}
The value of $ t $ used is 0.2 meters for Eigen and 0.1 meters for BTS and Plane-RCNN. 
.
We achieve precisions of 0.4965 and 0.4199, and recalls of
0.3585 and 0.4331 for the
under-estimated and over-estimated classes, respectively, when trained on the Eigen model,  as shown in Table \ref{table:dedn_single_view}. For
Plane-RCNN, we achieve a high precision of
0.5910 and a reasonable recall of 0.5250 for the under-estimated class.
We beat the random baseline on all 4 metrics for Plane-RCNN and 
3 out of 4 metrics for Eigen and BTS.

Eigen~\cite{eigen_iccv15} is the oldest of the three models used and hence has more distinct errors than the  newer models. Our results reflect this fact, as DEDN is able to classify  errors more accurately for Eigen than for the other two models (considering the average precision on both the error classes). BTS is the current state-of-the-art model and hence has the least distinct error patterns, which may explain why  our model cannot detect its errors nearly as accurately.



\subsubsection*{Multi-view Results}
For multi-view experiments, we used a value of $ t = 0.15$ metres for all three models. Table \ref{table:dedn_single_view} shows that we achieve a very high precision of 0.8862 and 0.8246 for
the under-estimated and over-estimated classes for Eigen. Even for BTS and Plane-RCNN, we achieve
high precisions of 0.5235 and 0.6934, and 0.6824 and 0.5538,
respectively, for the two error classes. Our results beat the random baseline by very large  margins 
on all of the metrics for Eigen and Plane-RCNN, and on 3 out of 4 metrics for BTS.
%
%
We note that the multi-view
results for each model are better than the corresponding results for
single-view, which supports our hypothesis that using multiple views
(two in this case) will aid the model to find errors more effectively.

Our architecture is very general and can be easily extended to more views for better accuracy.

\subsection{Correcting Errors from the Depth Prediction Models}

We try to iteratively improve the depth predictions using the predictions from our DEDN as explained in Section \ref{sec:error-correct-network}. 
We present results on 4 different widely-accepted error metrics: {Accuracy under a threshold ($\delta < thresh$), Absolute Relative Error (AbsRel), Root Mean Square Error (RMSE), and log10}; please refer to~\cite{lee_arxiv19,cadena_iros16} for definitions. 

\subsubsection*{Single-view Results}
Our DECN improved the depth map predictions on most of the
metrics for Eigen and Plane-RCNN (Table
\ref{table:iterative-improvement}). For Eigen and Plane-RCNN,  DECN improves the performance on all 4 
 metrics. 
%
%
DECN did not improve the BTS model, presumably due to the low precision and recall  for the  {single-view DEDN}.
%
%
We used 15 iterations for all the three models. 
We measured improvement in depth prediction in successive iterations but did not notice any improvement after the 15th iteration. 
Note that the ``Before" values are slightly different than the ones reported in the original papers since we have re-run their experiments. 


\begin{table}[t]
\begin{center}
\textbf{Single-view}
\end{center}\vspace{-4pt}
\resizebox{\columnwidth}{!}{
\begin{tabular}{lcc|ccc} \toprule
\multicolumn{2}{l}{\multirow{3}{*}{Model}} & \multicolumn{4}{c}{Metrics}                                                            \\ \cmidrule(lr){3-6}
\multicolumn{2}{c}{}                       & $\delta$ \ \textless \ 1.25 ($\uparrow$) & AbsRel ($\downarrow$) & RMSE ($\downarrow$) & $\log$ 10 ($\downarrow$) \\ \midrule
\multirow{2}{*}{Eigen~\cite{eigen_iccv15}}         & Before    &  0.6095    & 4.1154  & 0.8364  & 0.1233      \\
                               & After     &  \textbf{0.6096}    & \textbf{3.9943}   & \textbf{0.8297}   & \textbf{0.1224} \\ \\ 

\multirow{2}{*}{PlaneRCNN~\cite{Liu_CVPR19}}     & Before    &  0.8560    & 0.1260  & 0.2522    & 0.0544      \\
                               & After     &  \textbf{0.8655}    & \textbf{0.1233}  & \textbf{0.2403}  & \textbf{0.0523} \\ \\

\multirow{2}{*}{BTS~\cite{lee_arxiv19}}           & Before    &  \textbf{0.8958}   & \textbf{0.1071}   & \textbf{0.3853}   & \textbf{0.0454}      \\
                               & After     &  0.8916    & 0.1087   & 0.3989   & 0.0472 \\ \bottomrule
\end{tabular}
}
\begin{center}{\textbf{Multi-view}} \end{center} \vspace{-4pt}
\resizebox{\columnwidth}{!}{
\begin{tabular}{lcc|ccc} \toprule
\multicolumn{2}{l}{\multirow{3}{*}{Model}} & \multicolumn{4}{c}{Metrics}                                                            \\ \cmidrule(lr){3-6}
\cmidrule(lr){3-3} \cmidrule(lr){4-6}
\multicolumn{2}{c}{}                       & $\delta$ \ \textless \ 1.25 ($\uparrow$) & AbsRel ($\downarrow$) & RMSE ($\downarrow$) & $\log$ 10 ($\downarrow$) \\ \midrule
\multirow{2}{*}{Eigen~\cite{eigen_iccv15}}         & Before    &  0.4109    & \textbf{0.3518} & 1.0646   & 0.1447      \\
                               & After     &  \textbf{0.4329}    & 0.3706 & \textbf{1.0568}   & \textbf{0.1410} \\ \\ 
\multirow{2}{*}{PlaneRCNN~\cite{Liu_CVPR19}}     & Before    &  0.8477   & 0.1208 & 0.3319   & 0.0539      \\
                               & After     &  \textbf{0.8645}    & \textbf{0.1191} & \textbf{0.3035}  & \textbf{0.0497}      \\ \\
\multirow{2}{*}{BTS~\cite{lee_arxiv19}}           & Before    &  \textbf{0.8882}   & \textbf{0.1087}  & \textbf{0.3852}   & \textbf{0.0462}      \\
                               & After     &  0.8750   & 0.1151  & 0.4076   &  0.0513     \\ \bottomrule
\end{tabular}}
\caption{Accuracy of depth maps produced by single-view (top) and multi-view (bottom) DECN.}
\label{table:iterative-improvement}
\end{table}

\subsubsection*{Multi-view Results}

For the  multi-view case, DECN
improves on Eigen for 3 out of 4 metrics and on Plane-RCNN for all 4 metrics  (Table
\ref{table:iterative-improvement}). Our model does not improve for BTS, again probably because of the very low under-estimated recall achieved
for  multi-view DEDN for BTS.
%
We used 20 iterations for Eigen, 10 for BTS, and 15 for Plane-RCNN.

\subsection{Model Runtime}
Our model uses a ResNet50 U-Net. The running time of a single-view DEDN is 0.05 seconds (20 \textit{fps}). 
Single-view DECN requires about 0.76 per frame.
For the multi-view case, we pass  multiple views through the encoder sequentially, which adds  overhead and increases the DEDN  time
 to 
0.08 seconds (12.5 \textit{fps}) and DECN to 1.15 seconds.
We used an NVIDIA Titan Xp GPU for all 
experiments; a newer GPU would increase  speed significantly.

\section{CONCLUSIONS}
We present techniques for error diagnosis to analyze  monocular depth estimation models. The  \textit{Depth Error Detection Network (DEDN)}  locates  erroneous  depth pixels in single-view and multi-view settings. 
Our \textit{Depth Error Correction Network (DECN)}  improves predicted depth maps, and we show that it improves the results of Eigen and PlaneRCNN. 
Future work will evaluate using more than two views,
and  on outdoor scenes. 


\addtolength{\textheight}{-12cm}   




\noindent 
{\small{\textbf{Acknowledgements.}
 This work was supported by the Electronics and Telecommunications Research
 Institute (ETRI)  funded by the Korean government (21ZH1200, The
research of the fundamental media contents technologies for
hyper-realistic media space).}}


\bibliographystyle{IEEEtran}
\bibliography{IEEEfull}

\end{document}